\documentclass{article}
\usepackage{spconf,amsmath,graphicx,enumerate}
\usepackage[pdftex,linkcolor=black,citecolor=black,backref=page,
			colorlinks,
            linkcolor=red,
            anchorcolor=blue,
            citecolor=green,
            urlcolor=black,
            hyperfootnotes=true]{hyperref}          
\title{Computer vision-based food calorie estimation: dataset, method, and experiment}
\name{Yanchao Liang, Jianhua Li\thanks{This project is Supported by National Natural Science Foundation of China(No.61402174)}}
\address{
School of Information Science and Engineering, East China University of Science and Technology, China\,\\
\\\texttt{jhli@ecust.edu.cn}
}
\begin{document}
%\ninept
%
\maketitle
\begin{abstract}
Computer vision has been introduced to estimate calories from food images. But current food image datasets don’t contain volume and mass records of foods, which leads to an incomplete calorie estimation. In this paper, we present a novel food image dataset with volume and mass records of foods, and a deep learning method for food detection, to make a complete calorie estimation. Our dataset includes 2978 images, and every image contains corresponding each food’s annotation, volume and mass records, as well as a certain calibration reference. To estimate calorie of food in the proposed dataset, In this paper, we present a novel food image dataset with volume and mass records of foods. To estimate calorie of food in the proposed dataset, a deep learning method using Faster R-CNN is used to detect the food and calibration object
; GrabCut algorithm is used to get each food's contour. Then we estimate each food's volume and calorie. The experiment results show our estimation method is effective. Our dataset is the first released food image dataset, which can be used to evaluate computer vision-based calorie estimation methods. 
\end{abstract}

\section{Introduction}

Obesity is a medical condition in which excess body fat has accumulated to the extent that it may have a negative effect on health. People are generally considered obese when their Body Mass Index(BMI) is over 30 ${kg/m^2}$. High BMI is associated with the increased risk of diseases, such as heart disease, type two diabetes, etc\cite{BMI}. Unfortunately, more and more people will meet criteria for obesity\cite{柳叶刀肥胖}. 
The main cause of obesity is the imbalance between the amount of food intake and energy consumed by the individuals. Therefore, to lose weight in a healthy way, as well as to maintain a healthy weight for normal people, the daily food intake must be measured. However, current obesity treatment techniques require the patient to record all food intakes per day. In most of the cases, unfortunately patients have troubles in estimating the amount of food intake because of the self-denial of the problem, lack of nutritional information, the manual process of writing down this information (which is tiresome and can be forgotten), and other reasons.
Obesity treatment requires the patients to eat healthy food and decrease the amount of daily calorie intake, which needs patients to calculate and record calories from foods every day. While computer vision-based measurement methods were introduced to estimate calories from images directly according to the calibration object and foods information, obese patients have benefited a lot from these methods.

In recent years, there are a lot of methods based on computer vision proposed to estimate calories\cite{circle_plate,collaboration_card,ebutton,mobile_cloud}. Among these methods, the accuracy of estimation result is determined by two main factors: object detection algorithm and volume estimation method. In the aspect of object detection, classification algorithms like Support Vector Machine(SVM)\cite{SVM} are used to recognize food’s type in general conditions. In the aspect of volume estimation, the calibration of food and the volume calculation are two key issues. For example, when using a circle plate\cite{circle_plate} as a calibration object, it is detected by ellipse detection; and the volume of food is estimated by applying corresponding shape model. Another example is using people’s thumb as the calibration object, the thumb is detected by color space conversion\cite{RGB2YCBCR}, and the volume is estimated by simply treating the food as a column. However, thumb’s skin is not stable and it is not guaranteed that each person’s thumb can be detected. The involvement of human's assistance\cite{collaboration_card} can improve the accuracy of estimation but consumes more time, which is less useful for obesity treatment. After getting food’s volume, food's calorie is calculated by searching its density in food density table\cite{density_table} and energy in nutrition table\footnote{\url{http://www.hc-sc.gc.ca/fn-an/nutrition/fiche-nutri-data/nutrient_value-valeurs_nutritives-tc-tm-eng.php}}. Although these methods mentioned above have been used to estimate calories, the accuracy still need to be improved in the following two aspects: using widely-used calibration objects and more effective object detection algorithms.

Among the above measurement methods, a corresponding image dataset is in need, which is used to train and test the object detection algorithm. Several food image datasets\cite{food101,PFID,FooDD} have been created so far. A food dataset called Food-101 is proposed, which contains a lot of fast food images\cite{food101}. As those fast food images in Food-10 do not include a calibration object as a reference, estimating calorie is impossible in this dataset. Pittsburgh Fast-food Image Dataset(PFID)\cite{PFID} introduces a dataset including still images and videos of fast foods. The dataset named FOODD\cite{FooDD} comprises 3000 food images under different shooting conditions. Although those datasets can be used to train and test object detection algorithms, it is still hard to utilize them to estimate calories just because they have not prepared the volume and mass as a reference. There is still in need of Food image dataset for calorie estimation. 

  In this paper, a dataset named ECUST Food Dataset (ECUSTFD) and a novel calorie estimation method based on deep learning are proposed. In ECUSTFD, food’s volume and mass records are provided, as well as One RMB Yuan coin is used as the calibration object, which is widely used in daily life. For our calorie estimation method, it takes 2 images as its inputs: a top view and a side view of the food; each image includes a calibration object which is used to estimate image’s scale factor. Food(s) and calibration object are detected by object detection method called Faster R-CNN and each food’s counter is obtained by applying GrabCut algorithm. After that, we estimate each food's volume and calorie. 

The main contributions of this paper are listed as follows:
\begin{enumerate}[(1)]
 \item Proposing the first released image dataset with volume and mass records for food calorie estimation.
 \item  Proposing a complete and effective calorie estimation method.
\end{enumerate}

\section{DATABASE GENERATION}
\label{dateset_details}

\subsection{Dataset Description}
Nowadays, a corresponding food dataset is necessary for assessing those calorie estimation methods. That is why we create ECUSTFD.

\begin{figure*}[th]
\includegraphics[width=\textwidth]{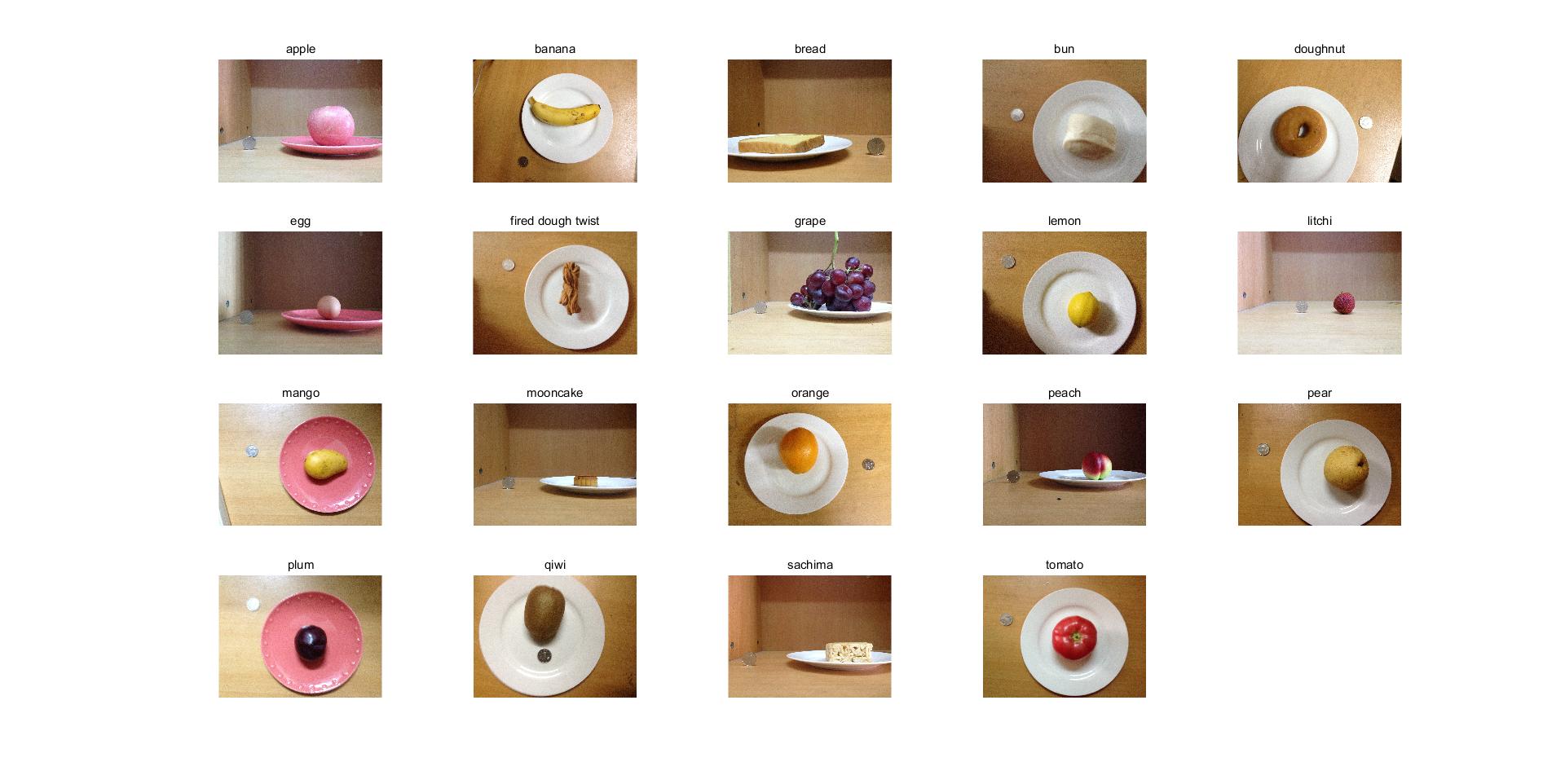}
\caption{ECUSTFD Sample Images}
\label{food samples}
\end{figure*}

As shown in Figure \ref{food samples},ECUSTFD contains 19 kinds of food: {\it apple, banana, bread, bun, doughnut, egg, fired dough nut, grape, lemon, litchi, mango, mooncake, orange, peach, pear, plum, qiwi, sachima, tomato}. The total number of food images is 2978. The number of images and the number of objects for the same type are shown in Table \ref{ECUSTFD table}. For a single food portion, we took several pairs of images by using smart phones; each group of images contains a top view and a side view of this food. For each image, there is only a One Yuan coin as calibration object and no more than two foods in it. If there are two food in the same image, the type of one food is different from another. We provide two datasets for researchers: one includes original images and another includes resized images. The size of each image in resized dataset is less than 1000${\times}$1000.

\begin{table}[th]
\caption{ECUSTFD}
\label{ECUSTFD table}
\centering
\addtolength{\tabcolsep}{-6pt}
{\begin{tabular}{|c|c|c|c|c|} %\toprule
\hline
Food Type & The number& The number & Density & Energy \\
& of images& of objects&($g/cm^3$)&($kcal/g$)\\
\hline
apple & 296 & 19 &0.78 &\hphantom{0}0.52 \\
banana & 178 & 15 &0.91 &\hphantom{0}0.89 \\
bread & \hphantom{0}66 & \hphantom{0}7 &0.18 &\hphantom{0}3.15 \\
bun & \hphantom{0}90 & \hphantom{0}8 &0.34 &\hphantom{0}2.23\\ 
doughnut & 210 & \hphantom{0}9 &0.31 &\hphantom{0}4.34\\ 
egg & 104 & \hphantom{0}7 &1.03 &\hphantom{0}1.43\\ 
fired dough twist & 124 & \hphantom{0}7 &0.58 &24.16\\ 
grape & \hphantom{0}58 & \hphantom{0}2 &0.97 &\hphantom{0}0.69\\ 
lemon & 148 & \hphantom{0}4 &0.96 &\hphantom{0}0.29\\ 
litchi & \hphantom{0}78 & \hphantom{0}5 &1.00 &\hphantom{0}0.66\\ 
mango & 220 & 10 &1.07 &\hphantom{0}0.60\\ 
mix & 108 & 14 &/ &/\\ 
mooncake & 134 & \hphantom{0}6 &0.96 &18.83\\ 
orange & 254 & 15 &0.90 &\hphantom{0}0.63\\ 
peach & 126 & \hphantom{0}5 &0.96 &\hphantom{0}0.57\\ 
pear & 166 & \hphantom{0}6 &1.02 &\hphantom{0}0.39\\ 
plum & 176 & \hphantom{0}4 &1.01 &\hphantom{0}0.46\\ 
qiwi & 120 & \hphantom{0}8 &0.97 &\hphantom{0}0.61\\ 
sachima & 150 & \hphantom{0}5 &0.22 &21.45\\ 
tomato & 172 & \hphantom{0}4 &0.98 &\hphantom{0}0.27\\ 
\hline
%\botrule
\end{tabular}}
\end{table}
For each images, our dataset still provide other informations as follows:
\begin{enumerate}[(1)]
\item \textbf{Annotation}. Our dataset still provide bounding boxes (we only annotated the resized images because the original images are more than screen resolution and is hard for us to annotate) for each object in every images. For example,  we offer two bounding boxes for a single apple image: one for the apple and another for the calibration object.
\item \textbf{Mass}. We provide mass for each food. The mass is obtained with an electronic scale.
\item \textbf{Volume}. Considering that we can only get volume from food images rather than mass, we choose to provide the volume information as a reference. The volume is measured with drainage method. Due to the limit of containing cup, the volumes we measured in ECUSTFD are not as reliable as the qualities we measured.
\item \textbf{Density and Energy}. In order to estimate calorie, foods' density and energy information should be provided. The density is calculated with the volume and mass information collected in ECUSTFD. For each kind of food, energy is obtained from nutrition table. 
\end{enumerate}

\subsection{Shooting Conditions}
We took into consideration important factors that affect the accuracy of estimation results: camera, lighting, shooting angle, displacement, calibration object, food type. 
\begin{enumerate}[(1)]
\item \textbf{Camera}. We use iPhone 4s and iPhone 7 to take photos. For the same scene, the images taken from different cameras may be different from each other due to the performances of cameras and algorithms. For most of images in ECUSTFD, the size of image taken by iPhone 4s is 2448${\times}$3264 and the size of image taken by iPhone 7 is 3024${\times}$4032.
\item \textbf{Lighting}. As people can eat food on the table anytime, the photos in our dataset are taken from different lighting conditions. Some photos are taken in dark environment with or without flash light. 
\item \textbf{Shooting angles}. When taking a top view, shooting angle is almost 0 degree from the table; and when taking a side view, shooting angle is almost 90 degree from the table. 
\item \textbf{Displacement}. For a food image in our dataset, the position of food is not fixed. It means that food can be placed in anywhere as long as this food can be captured completely by camera. So as the calibration object. In most cases, the food is put on a red or white plate; in other cases, food is on the dining table directly.
\item \textbf{Calibration object}. We choose One Yuan coin as this dataset's Calibration object, which is easy to get in our daily life. The diameter of One Yuan coin is 25.0${mm}$ as shown in Figure \ref{One Yuan coin}. One Yuan coin can be detected by Hough Transform\cite{houghcircle} or deep learning methods.
\begin{figure}[htbp]
\centering{\includegraphics[width=4.5cm]{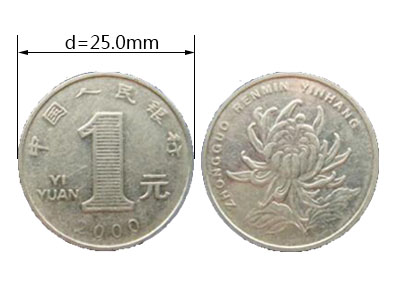}}
\caption{Two Sides of One Yuan Coin}
\label{One Yuan coin}
\end{figure}
\item \textbf{Food type}. For obtaining food’s volume and mass easily, we only choose those foods which are big enough, stable and less prone to deformation. If food with small volume, like peanut, is hard to get its volume and will cause great error when comparing with its real volume. Every food in our dataset is complete. We prefer to use a whole apple to take photos rather than sliced apple, which makes it easy to measure volume and weight. In reality, the calorie of an apple with skin is higher than the calorie of the same apple without skin.
\end{enumerate}

\subsection{ACCESSMENT}
ECUSTFD is a free public food image dataset. The dataset with original images and no annotations is publicly available at this website\footnote{\url{http://pan.baidu.com/s/1dF866Ut}}. The small image dataset including annotations, volume and mass information is available at this website\footnote{\url{https://github.com/Liang-yc/ECUSTFD-resized-} or \url{http://pan.baidu.com/s/1o8qDnXC}}. You will find instructions at that websites either.
\section{CALORIE ESTIMATION METHOD}
\label{system}
\subsection{Calorie Estimation Method Based On Deep Learning}
Before performing the experimental results, we briefly introduce our calorie estimation method. 

\begin{figure}[th]
\centering
\includegraphics[width=5.5cm]{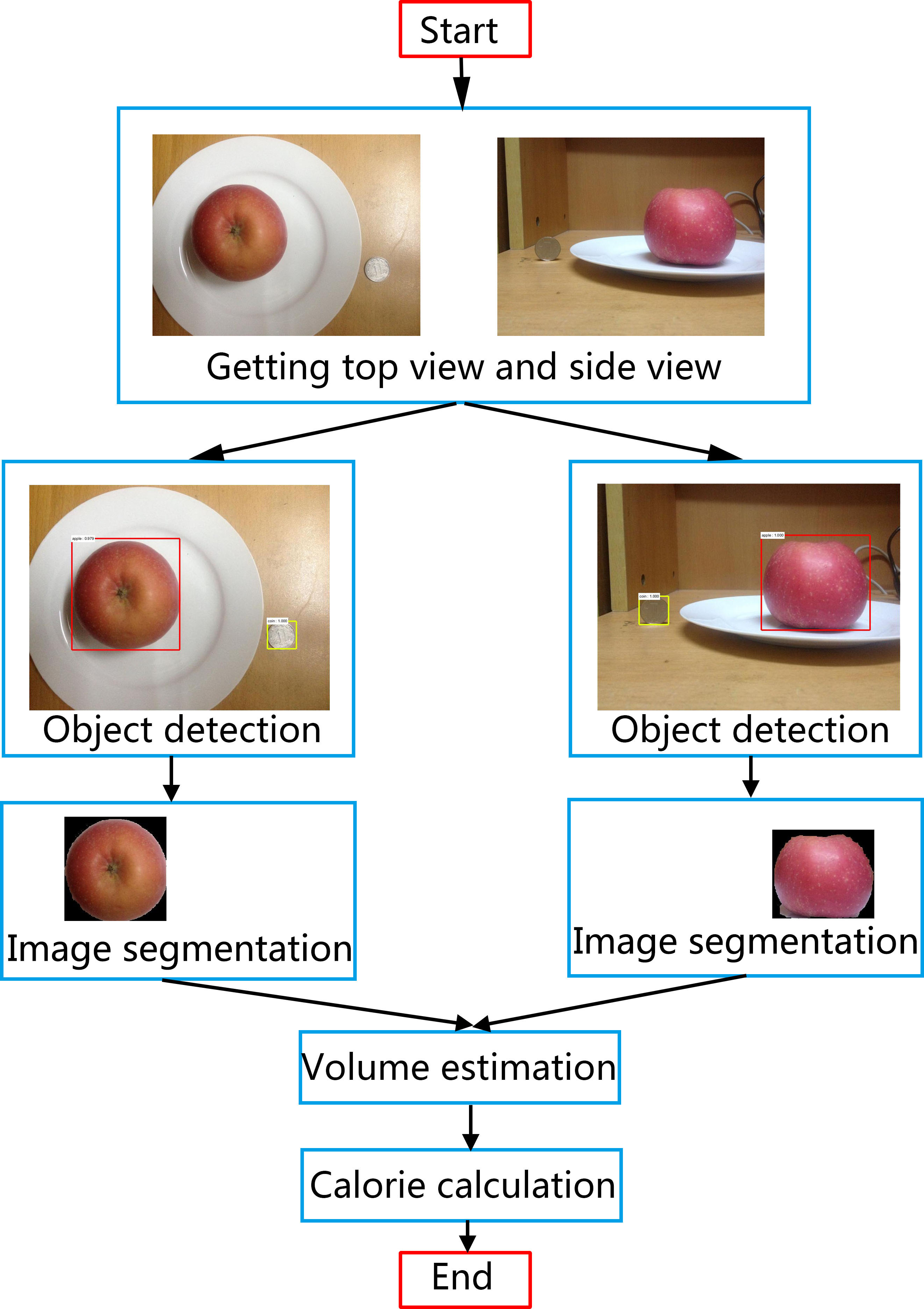}
\caption{Calorie Estimation Flowchart}
\label{calorie estimation system}
\end{figure}

Our goal is to help obese people to calculate the calories they get from the food. Figure \ref{calorie estimation system} shows the flowchart of the proposed system. To estimate calories, it requires the user to take a top view and a side view of the food before eating with his/her smart phone. Each images used to estimate must include One Yuan coin. For the top view, we use the deep learning algorithms to recognize the types of food and apply image segmentation to identify the food's contour in the photos. So as the side view. then, the volumes of each food is calculated based on the calibration objects in the images. In the end, the calorie of each food is obtained by searching density table and nutrition table.
In order to get better results, we choose to use Faster Region-based Convolutional Neural Networks (Faster R-CNN)\cite{fasterrcnn} to detect objects and GrabCut \cite{grabcut} as segmentation algorithms. 
\subsection{Objection detection With Deep Learning Methods}
We do not use semantic segmentation method such as Fully Convolutional Networks (FCN)\cite{fcn} but choose to use Faster R-CNN. Faster R-CNN is a framework based on deep convolutional networks. It includes a Region Proposal Network (RPN) and an Object Detection Network\cite{fasterrcnn}. When we put an image with RGB channels as input, we will get a series of bounding boxes. For each bounding box created by Faster R-CNN, its class is judged. 
\subsection{Image Segmentation}
Before estimating volume, we choose to segment each bounding box first. GrabCut is an image processing approach based on optimization by graph cuts\cite{grabcut}. Practicing GrabCut needs user to draw a bounding box around the object; and such boxes can be provided by Faster R-CNN. Although asking user to label the foreground/background color can get better result, we refuse it so that our system can finish calorie estimation without user’s assistance. For each bounding box, we get precious contour after applying GrabCut algorithm. Then we can estimate every food’s volume and calorie.
\subsection{Volume Estimation And Calorie Calculation}
According to the One Yuan coin detected in the top view, the true size of a pixel is known. Similarly, we know actual size of of a pixel in the side view. Then we use different formulas to estimate volume of each food. After getting volume, food's calorie is obtained by searching related tables.
\section{EXPERIMENT}
\label{experiment}
In this section, we present the volume estimation results using the food images dataset. These food and fruit images are divided into train and test sets. In order to avoid using train images to estimate volumes, the images of two sets are not selected randomly but orderly. 
\begin{figure*}[th]
\centering
\includegraphics[width=\textwidth]{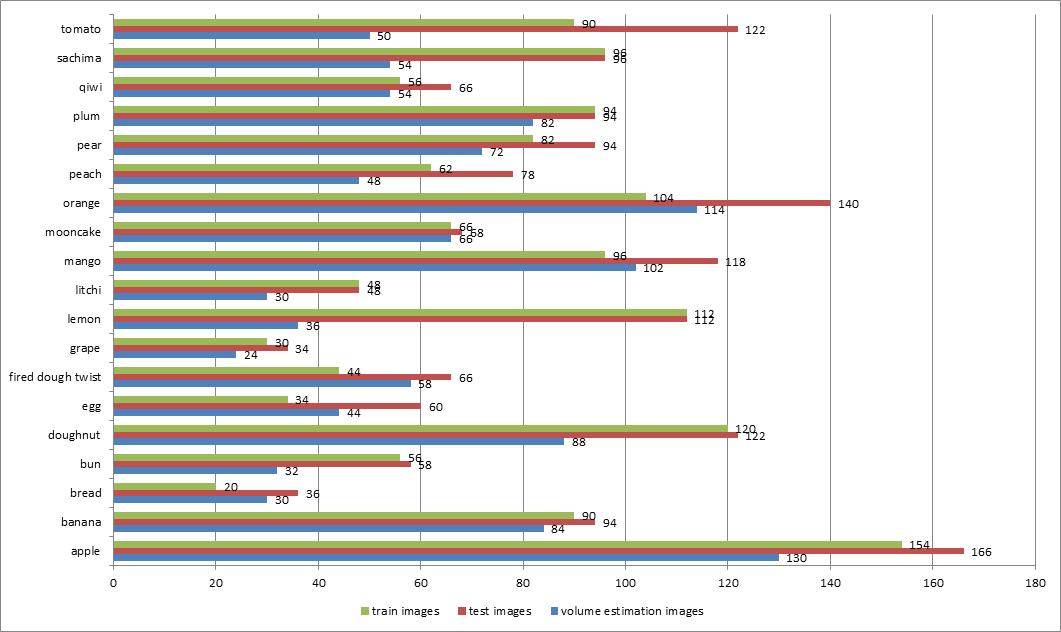}
\caption{Image Number in Experiment}
\label{image number}
\end{figure*}
The numbers of train and test images used for Faster R-CNN are listed in Figure \ref{image number}. After Faster R-CNN is well trained,  we use those pairs of test images which Faster R-CNN correctly recognizes to estimate volumes. In other words, those images Faster R-CNN cannot identity or misidentify in test sets will be discarded. The numbers of images in volume estimation experiments are shown in Figure \ref{image number} either. The code can be downloaded at this website\footnote{\url{https://github.com/Liang-yc/CalorieEstimation}}. We use mean error to evaluate volume estimation results. Mean error {\it ME} is defined as:
\begin{equation}
\label{ME}
{\it ME_i}=\frac{1}{n_i} \sum_{j=1}^{n_i} \frac{v_j-\uppercase{V}_j}{\uppercase{V}_j}
\end{equation}
In Equation \ref{ME}, for food type ${i}$ , $2{n_i}$ is the number of images Faster R-CNN recognizes correctly. Since we use two images to calculate volume, so the number of estimation volumes  for ${i}$th  type is ${n_i}$. ${v_j}$ is the estimation volume for the ${j}$th pair of images with the food type ${i}$; and ${\uppercase{V}_j}$ is corresponding estimation volume for the same pair of images.

\begin{figure*}[htbp]
\centering
\includegraphics[width=\textwidth]{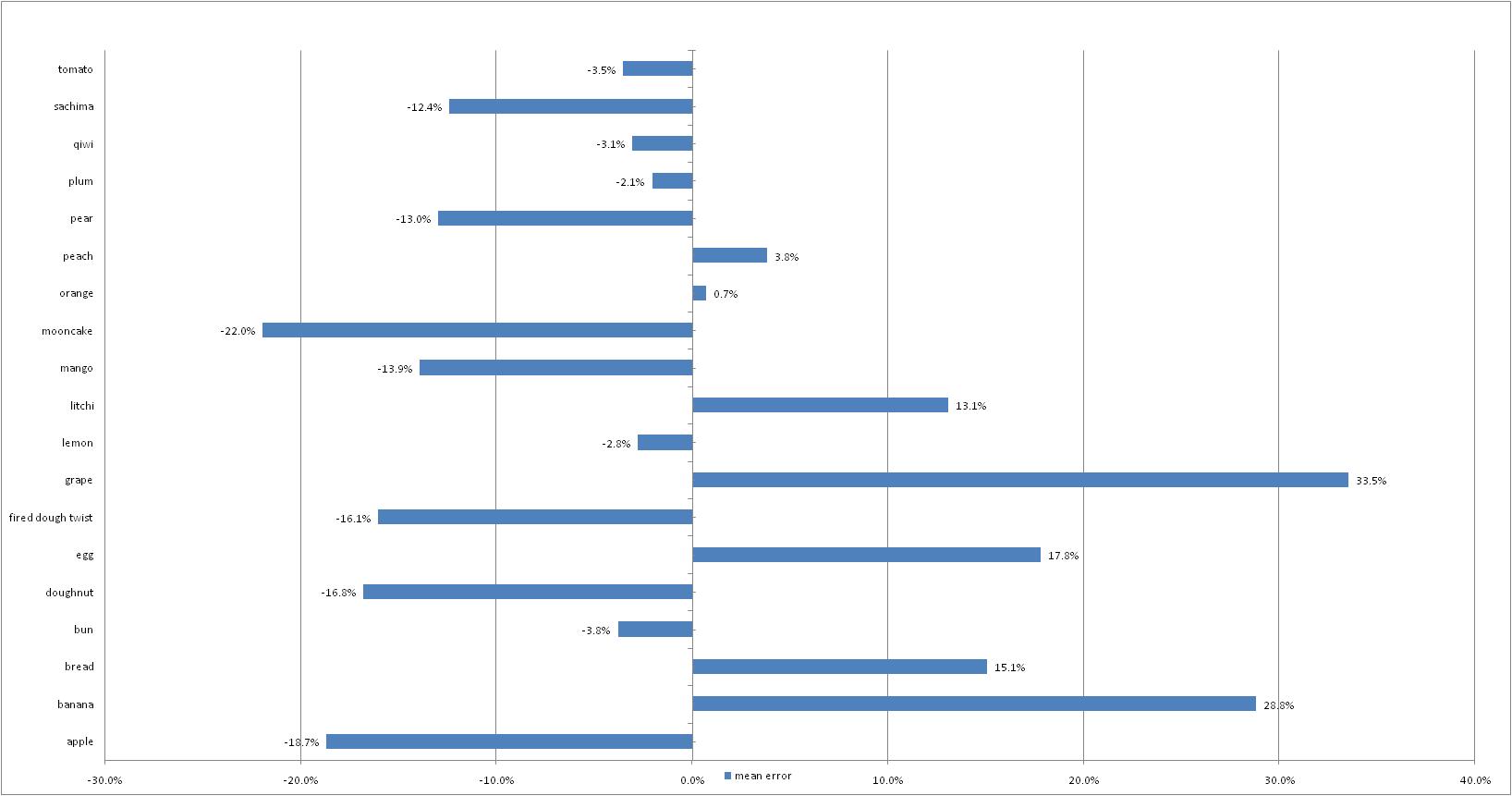}
\caption{Volume Estimation Results}
\label{volume error}
\end{figure*}

Volume estimation results are shown in Figure \ref{volume error}. For most types of food in our experiment, the estimation volume are closer to reference volume. The mean error between estimation volume and true volume does not exceed ${\pm}$20\% except banana, grape, mooncake. For some food types such as orange, our estimation result is close enough to the true value. As for the bad estimation results of grape, the way we measure volumes of grapes should be to blame. When measuring grapes' volumes, we have not used plastic wrap to cover grapes but put them into the water directly, so the volumes we estimated are far from the reference values because of the gaps. All in all, our estimation method is available.
\section{CONCLUSION}
\label{conclusion}
In this paper, we provided a dataset called ECUSTFD. A main feature of ECUSTFD is that each food’s volume and mass records are provided. We provided experimental results with deep learning algorithms in ECUSTFD.
\bibliographystyle{IEEEbib}
\bibliographystyle{unsrt}
\bibliography{refs}
\end{document}